\pdfoutput=1

\documentclass[11pt]{article}

\usepackage[final]{acl}

\usepackage{svg}
\usepackage{times}
\usepackage{latexsym}
\usepackage{hyperref}
\usepackage{pgfplots}
\usepackage{pgfplotstable}
\usepgfplotslibrary{fillbetween}
\usepackage{enumitem}
\usepackage{graphics}
\usepackage{enumitem}
\usepackage{makecell}
\usepackage{graphicx}
\usepackage{fvextra}
\usepackage{spverbatim}
\usepackage{xcolor}
\usepackage{booktabs}
\usepackage{multirow}
\usepackage[framemethod=TikZ]{mdframed}
\usepackage[normalem]{ulem}
\useunder{\uline}{\ul}{}

\usepackage{amsfonts}

\usepackage{subfigure}

\usepackage{booktabs}
\usepackage{multirow}
\usepackage[normalem]{ulem}
\useunder{\uline}{\ul}{}

\usepackage{algorithm}
\usepackage{algorithmic}

\usepackage{amsmath}
\usepackage{amsfonts}
\usepackage{caption}

\usepackage[T1]{fontenc}

\usepackage[utf8]{inputenc}

\usepackage{microtype}

\usepackage{inconsolata}

\usepackage{graphicx}

\renewenvironment{itemize}{
    \begin{list}{\labelitemi}{
    \setlength{\topsep}{5pt}
    \setlength{\partopsep}{5pt}
    \setlength{\parskip}{3pt}
    \setlength{\itemsep}{0pt}
    \setlength{\parsep}{0pt}
    }
}{\end{list}}

\mdfsetup{%
   middlelinecolor=black,
   middlelinewidth=1pt,
   backgroundcolor=gray!10,
   roundcorner=5pt,
   frametitlerule=true
}

\title{Through the Looking Glass: Common Sense\\Consistency Evaluation of Weird Images}

\author{
\bf Elisei Rykov\textsuperscript{1}\quad
\bf Kseniia Petrushina\textsuperscript{1,4}\quad
\bf Kseniia Titova\textsuperscript{1,3}\\
\bf Anton Razzhigaev\textsuperscript{2}\quad
\bf Alexander Panchenko\textsuperscript{1,2}\quad
\bf Vasily Konovalov\textsuperscript{2,4}\\
\textsuperscript{1}Skoltech\quad 
\textsuperscript{2}AIRI\quad
\textsuperscript{3}MTS AI\\
\textsuperscript{4}Moscow Institute of Physics and Technology\\
\{\href{mailto:Elisei.Rykov@skol.tech}{Elisei.Rykov}, \href{mailto:Kseniia.Petrushina@skol.tech}{Kseniia.Petrushina}, \href{mailto:A.Panchenko@skol.tech}{A.Panchenko}\}@skol.tech
}

\begin{document}
\maketitle
\begin{abstract}

Measuring how real images look is a complex task in artificial intelligence research. For example, an image of a boy with a vacuum cleaner in a desert violates common sense. We introduce a novel method, which we call Through the Looking Glass (TLG), to assess image common sense consistency using Large Vision-Language Models (LVLMs) and Transformer-based encoder. 

By leveraging LVLMs to extract atomic facts from these images, we obtain a mix of accurate facts. We proceed by fine-tuning a compact attention-pooling classifier over encoded atomic facts. Our TLG has achieved a new state-of-the-art performance on the WHOOPS! and WEIRD datasets while leveraging a compact fine-tuning component.\footnote{\url{https://github.com/s-nlp/through-the-looking-glass}} 

\end{abstract}

\section{Introduction}
People quickly notice something unusual in images that defy common sense, like Einstein holding a smartphone. We find it odd even though each part seems normal. Our brain's ability to understand normality goes beyond just identifying objects~\cite{zellers2019recognition}. It involves connecting visual cues with everyday knowledge.

In this work, we propose a visual commonsense model that utilizes the observation that LVLMs may generate contradictory facts when confronted with images defying common sense~\cite{phd}. By leveraging LVLMs to extract atomic facts from these images, we obtain a mix of accurate facts and erroneous hallucinations. Then we fine-tune a compact attention-pooling model over encoded atomic facts.

Our results indicate that using the classifier for basic facts can efficiently spot strange images. Surprisingly, this method outperforms existing more complex techniques. 

In addition, we introduce a synthesized WEIRD dataset, a dataset of 824 samples of normal and strange images. Using this dataset, we
further confirmed the performance of our model.

Our contributions are as follows:
\begin{itemize}
\item We present a \textit{new method} called TLG that achieved state-of-the-art performance on the existing dataset of normal and strange images WHOOPS!.
\item We present a \textit{new dataset} dubbed WEIRD which is more challenging and 
nearly four times larger than WHOOPS!. 
\end{itemize}

\section{Related Work}

Recently, commonsense reasoning has attracted substantial interest from the research community, spanning disciplines within NLP and CV, with numerous tasks being introduced. 

~\citet{whoops} introduced the WHOOPS! benchmark, comprised of purposefully commonsense-defying images created by designers using publicly available image generation tools like Midjourney. They used a supervised approach based on BLIP-2 Flan-T5~\cite{li2023blip2} on multiple scales. The proposed fine-tuned model managed to outperform a random baseline, but still falls significantly short of human performance.

\begin{figure*}[!ht]
\centering
\includegraphics[width=0.99\textwidth]{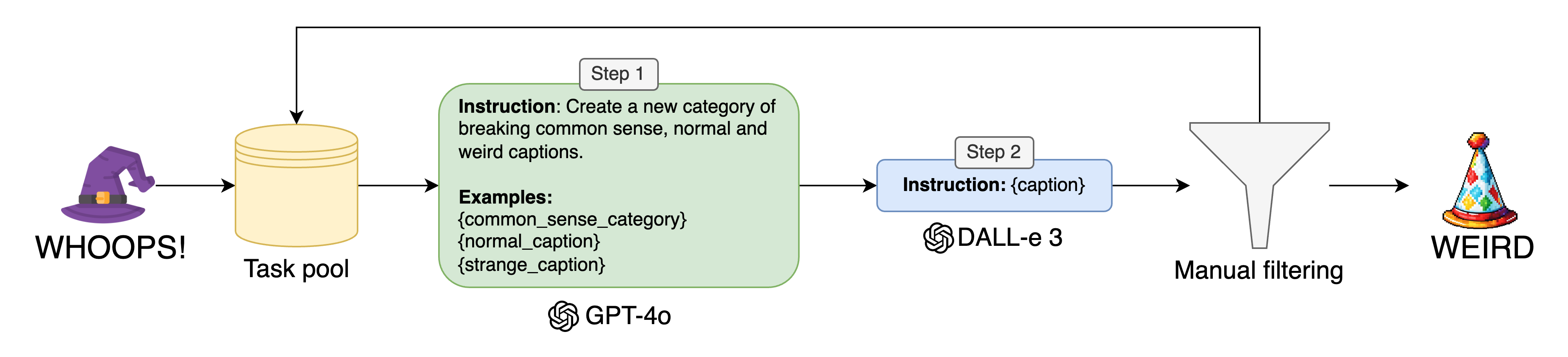}
\caption{WEIRD dataset generation process. First, we formed a task pool for the few-shot generation of new samples from the WHOOPS! benchmark. Next, we randomly sampled few-shots from the task pool and asked GPT-4o to generate new samples. The samples were then visualized using Dall-E 3 and manually filtered. Good samples were added to the task pool for the next few-shot sampling. 
}
\label{weird_schema}
\end{figure*}

LLMs are capable of producing highly fluent responses to a wide range of user prompts, but they are notorious for hallucinating and making non-factual statements. \citet{selfcheckgpt} proposed SelfCheckGPT, a straightforward sampling-based method that enables fact-checking of black-box models with zero resources. 

To assess consistency among multiple sampled responses, SelfCheckGPT utilizes several techniques, including BERTScore, an automatic multiple-choice question answering generation (MQAG) framework~\cite{manakul}, and NLI contradiction scores to detect hallucinations in the generated responses. However, the most effective method found was prompting the LLM to verify if the generations are supported by the context or not.

Regarding multi-modal case, \citet{faithscore} proposed FAITHSCORE, a reference-free and fine-grained evaluation metric that measures the faithfulness of the generated free-form answers from large vision-language models. The FAITHSCORE uses multistep approach: (1) identify the descriptive content, (2) extract corresponding atomic facts from the identified sentences, and (3) the faithfulness of all atomic facts is verified according to the input image by applying Visual Entailment Model (VEM), which is able to predict whether the image semantically entails the text. Analogously, NLI has been used in textual mode to verify premises and hypotheses and subsequently to detect hallucinations~\cite{maksimov-etal-2024-deeppavlov}.

~\citet{aaai} proposed an approach, in which LVLM is used to first generate atomic facts from images, resulting in a combination of accurate facts and erroneous hallucinations. The next step involves calculating pairwise entailment scores among these facts and aggregating these values to produce a single reality score.

Our approach is similar to the preceding methods, as we also utilize LVLMs to extract atomic facts from the image. We then train a supervised model to learn the relationships between the derived facts. If the classifier identifies a high contradiction among atomic facts, it indicates that one of the generated atomic facts is likely a hallucination. This often occurs when the LVLMs encounter an unusual image~\cite{phd}, leading to such inconsistencies in most cases.

\begin{table}[hbt!]
\centering
\small
\begin{tabular}{@{}lll@{}}
\toprule
                                        & \textbf{WHOOPS!} & \textbf{WEIRD} \\ \midrule
\# of samples                           & 204              & 824            \\
\# of categories & 26               & 12             \\
\# of sub-categories & ––               & 181             \\
Human baseline                          & 92\%             & 82.22\%        \\ \bottomrule
\end{tabular}
\caption{Comparison details between WHOOPS! and WEIRD. WEIRD contains 4 times more samples than WHOOPS!. In addition, WEIRD contains 181 different generated commonsense-breaking categories, which have been grouped into 12 global categories.}
\label{tab:whoops_weird_comp}
\end{table}


\section{Dataset}
This section describes the datasets we used to evaluate our methodology.

\subsection{WHOOPS!}
To evaluate our methods, we employ the WHOOPS!\footnote{\textbf{W}eird and \textbf{H}eterogene\textbf{O}us \textbf{O}bjects, \textbf{P}henomena, and \textbf{S}ituations} benchmark, focusing on a subset comprising 102 pairs of weird and normal images. Performance is measured by binary accuracy within this paired dataset, where a random guess would yield 50\% accuracy. To assess human performance, three annotators were enlisted to categorize each image as weird or normal, relying on a majority vote for the final determination. Impressively,the human baseline reached 92\%, indicating that despite subjectivity, there is a clear consensus on what constitutes weirdness within the specific context of the WHOOPS! benchmark.

\begin{figure*}[!hbt]
  \centering
        \centering
        \small
        \includegraphics[width=1\textwidth]{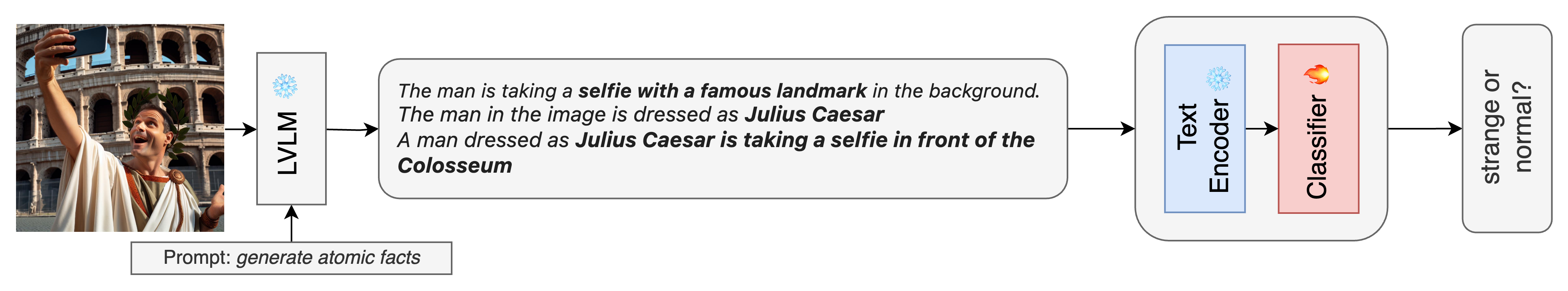}
        \caption{The proposed approach TLG for image commonsense consistency evaluation. Using the LVLM-generated atomic facts about the image, we train a classifier using hidden states from the textual encoder.}
        \label{fig:lvlm-based}
\end{figure*}

\subsection{WEIRD}
Due to the fact that the WHOOPS! benchmark is relatively small, we generated a larger benchmark for quantifying image realism to validate our methodology -- WEIRD\footnote{\textbf{W}eird \textbf{E}xamples of \textbf{I}mages with \textbf{R}eal-life \textbf{D}iscrepancies}. 

The detailed process of WEIRD dataset creation is shown in Figure~\ref{weird_schema}. Like the Self-Instruct~\cite{wang-etal-2023-self-instruct} dataset, WEIRD was generated in an iterative, semi-automatic manner using LLM. Specifically, we used WHOOPS! as an initial task pool with few-shot samples. In each iteration, we randomly sampled 5 pairs of normal and weird situations, along with the commonsense-breaking category. Each few-shot sample contains the breaking commonsense category, a caption of the normal image, and a caption of a strange image. The randomly sampled few-shots were passed to GPT-4o to generate a new category and captions. See the exact prompt used for generation in Appendix~\ref{app:weird-prompt}. In the next step, these textual captions were used to generate images with Dall-E 3.

In each iteration, we generated 50 pairs of normal and strange images, resulting in 100 samples after each iteration. We also manually filtered out bad samples. We considered bad samples to be those with inconsistencies between image and caption, or with textual noisy captions. For example, there were many inconsistencies in the captions that mention celebrities. It turned out that Dall-E 3 struggled with the generation of celebrity faces, while some strange captions were based on putting certain celebrities in inappropriate conditions.

In total, we generated 2,000 unique samples of commonsense-breaking situations before the filtering stage. After filtering, only 824 samples remained. %
To evaluate human performance on WEIRD, we additionally annotated the dataset on the Yandex Tasks\footnote{\url{https://tasks.yandex.com}} crowd-source platform. Each example was annotated by five annotators with overlapping assignments. In order to introduce crowd sources to the task, we added 10 training samples. As a result of the annotation process, Krippendorff's alpha coefficient of consistency was 0.69 with a human accuracy of 82.22\%. WHOOPS! and WEIRD comparison details can be seen in Table~\ref{tab:whoops_weird_comp}.

\section{Visual Commonsense Evaluation Method using Atomic Fact Extraction}

The idea of our method dubbed TLG (Through the Looking Glass) is inspired by FactScore \cite{min-etal-2023-factscore}: we adopt the principle of atomic facts generation for trustworthiness verification for the image modality. Namely, the common sense evaluation method is based on the classification of atomic facts generated by LVLMs using textual encoders. The approach is depicted in Figure~\ref{fig:lvlm-based}.

We use LVLMs to collect different atomic facts that describe different aspects of the scene in the image. To sample as many different facts as possible, we use the Diverse Beam Search~\cite{DBLP:journals/corr/VijayakumarCSSL16}. So, given an image $I$ and an LVLM, we sample N facts $F=\{ f_1, f_2, \dots, f_N \}$, where $F = \text{LVLM}(I)$.

Next, we use a frozen textual encoder to extract representations $H$ of the generated atomic facts. Each fact representation is computed as
\begin{equation}
    H_i = \text{Encoder}(f_i) \in \mathbb{R}^{N \times T \times d},
\end{equation}
where $T$ -- number of tokens, $d$ -- embeddings dimensionality.

Since each encoder output $H$ is a set of hidden representations for each token and fact, we perform average pooling to extract a single representation $V$ for each fact. Thus, using the attention masks $m$ obtained by the encoder tokenizer and the hidden representations $H$, we compute a single fact representation by averaging the vectors of its tokens
\begin{equation}
    V_i = \frac{\sum_{j=1}^{T} m_{ij} H_{ij}}{\sum_{j=1}^{T} m_{ij} + \varepsilon}.
\end{equation}

Furthermore, we train an attention-based pooling classifier using individual representations $V$. This classifier maps each representation to a single value. Then, we convert a set of attention values into probabilities using the softmax function:
\begin{equation}
    A = \text{softmax}(W_a V + b_a) \in \mathbb{R}^{N}.
\end{equation}

Later, these scores are used to perform a weighted averaging of the set of representations for each fact into a single representation:

\begin{equation}
    v_{\text{weighted}} = \frac{\sum_{i=1}^{N} A_i V_i}{\sum_{i=1}^{N} A_i} \in \mathbb{R}^{d}.
\end{equation}

Finally, we classify the final representation by mapping it to a single common sense violation probability:
 \begin{equation}
     \text{prob} = \sigma(W_c v_{\text{weighted}} + b_c) \in [0,1].
 \end{equation}

\begin{table*}[!htb]
\small
\centering
\begin{tabular}{@{}lcccc@{}}
\toprule
\textbf{Method}          & \textbf{\# Total} & \textbf{Mode}               & \textbf{WHOOPS!} & \textbf{WEIRD} \\ \midrule
Humans                  & --                & --                          & 92.00               & 82.22          \\ \midrule
BLIP2 FlanT5-XL         & 3.94B             & \multirow{2}{*}{fine-tuned} & 60.00            & 71.47          \\
BLIP2 FlanT5-XXL        & 12.4B             &                             & 73.00            & 72.31          \\ \midrule
BLIP2 FlanT5-XXL        & 12.4B             & \multirow{7}{*}{zero-shot}  & 50.00            & 63.84          \\
nanoLLaVA Qwen1.5 0.5B  & 1.05B             &                             & 66.66            & 70.90          \\
LLaVA 1.6 Mistral 7B    & 7.57B             &                             & 56.86            & 61.18          \\
LLaVA 1.6 Vicuna 7B     & 7.06B             &                             & 65.68            & 76.54          \\
LLaVA 1.6 Vicuna 13B    & 13.4B             &                             & 56.37            & 58.36          \\
InstructBLIP Vicuna 7B  & 7B                &                             & 61.27            & 69.41          \\
InstructBLIP Vicuna 13B & 13B               &                             & 62.24            & 66.58          \\ 
GigaChat-Pro & 30B               &                             & 65.19            & 71.62          \\ 

\midrule
Qwen2.5 7B Instruct     & 15.18B            & \multirow{2}{*}{zero-shot}  & 67.65            & 66.46          \\
Gemma2-9B               & 16.57B            &                             & 73.04            & 82.92          \\
\midrule
LP - LLaVA                & 13B                & fine-tuned & {\ul 73.50}            & {\ul 85.26}          \\
CLIP                    & 0.65B             &            --                 & 60.78            & 81.57          \\
TLG (Ours)           & 8B             &             fine-tuned                & \textbf{73.54}   & \textbf{87.57} \\ \midrule \midrule
GPT-4o                  & --                & zero-shot                   & 79.90            & 81.64          \\ \bottomrule
\end{tabular}
\caption{The results of different approaches on WHOOPS! and WEIRD datasets. Both benchmarks are balanced and accuracy is the evaluation metric. Fine-tuned methods are displayed at the top, while zero-shot methods are presented in the middle. The best linear probing results for all configurations along with our method are displayed at the bottom. }
\label{tab:performance}
\end{table*}

\vspace{-14pt}
\section{Experimental Setup}
\label{section:exp-setup}

To run the experiments, we strictly follow the evaluation setup suggested in WHOOPS!~\cite{whoops}. Thus, we evaluate several models using 5-fold cross-validation in a supervised configuration.  See the detailed list of checkpoints used for the main approach and baselines in Appendix~\ref{app:checkpoints}.

For fact generation, we set \texttt{num\_beams} and \texttt{num\_beam\_groups} to $5$, and the \texttt{diversity\_penalty} to $1.0$. Regarding penalty, we find this value to be optimal for adding diversity and preserving the model's ability to follow instructions. For LVLMs, with various backbone architectures, we utilized the following prompt for fact generation: \textit{``Provide a brief, one-sentence descriptive fact about this image''}. To generate atomic facts, we used different LVLMs with different sizes (from 0.5B to 13B) of the LLaVA architecture. Given the generated atomic facts, we encode them using several DeBERTa-v3-large-based encoders.

We also consider the following baselines:

\paragraph{LVLM} with the prompt, which was found to be effective in detecting weird images~\cite{liu2024improvedbaselinesvisualinstruction}: \textit{``<image> Is this unusual? Please explain briefly with a short sentence.''} 

\paragraph{Linear Probing} resemble our approach in that it requires a small learnable component. This baseline involves learning a logistic regression classifier on the hidden representation of LLaVAs at each layer. We consider two setups: (a) using the <image> as the sole input (\textbf{Image only}), and (b) using <image> the with a prompt \textit{``Provide a short, one-sentence descriptive fact about this image''} (\textbf{+Prompt}), which was used to generate atomic facts.

\paragraph{CLIP-based} models were evaluated by passing images and measuring the distance from the \texttt{strange} and \texttt{normal} classes in a zero-shot setting. In addition, we fine-tuned CLIP in a cross-validation setting. More details on the hyperparameters and detailed baseline results can be found in the Appendix~\ref{app:clip:baseline}. 
\vspace{-4pt}
\paragraph{LLM} zero-shot baselines were represented by Gemma-2-9B-Instruct and Qwen2.5-7B-Instruct. As input, we passed generated atomic facts about the image and asked the model to determine whether the facts were strange or not using the following prompt: \textit{``Your task is to classify a series of facts as normal or strange. The set of facts is strange if some of the facts contradict common sense. Answer using 'normal' or 'strange'. Do not write anything else''}. 

Furthermore, we used two fine-tuned baselines based on BLIP2~\cite{blip2}:  BLIP2 FlanT5-XL and BLIP2 FlanT5-XXL that were reported in~\citet{whoops}.

Moreover, we conducted experiments on knowledge transfer between WEIRD and WHOOPS! for fine-tunable methods to explore the generalization ability to another dataset.

\section{Results}

The results of our experiments on both WHOOPS! and WEIRD datasets are presented in Table~\ref{tab:performance}. The proprietary GPT-4o model is included as a baseline to illustrate the complexity of benchmarks for proprietary systems and to demonstrate the performance gap between human-generated and proprietary systems. It is not directly comparable to other open-source methods. The results of the linear probing baselines can be found in the Appendix~\ref{app:linear:probing}. For the TLG method and LLM-based baselines, we used facts produced by LLaVA 1.6 Mistral 7B; see the Appendix~\ref{app:text-encoder} for more details. The total number of parameters represents the sum of all parameters in the method. As LLMs and text encoders use pre-generated atomic facts, we report their parameters together with the LVLMs parameters. See Appendix~\ref{ref:analysis} for an analysis of generated facts.

\noindent \textbf{TLG} achieves an accuracy of 73.54\% on WHOOPS! and 87.57\% on WEIRD, demonstrating the state-of-the-art performance on both datasets.

\noindent \textbf{BLIP2 FlanT5 vs. TLG}
Next, we compare our approach to the baselines from~\citet{whoops}. TLG outperforms the original fine-tuned approach (BLIP2-FLAN-T5-XXL). This suggests that the task of detecting anomalous images should be tackled by fine-tuning a compact classifier on either textual representations or images, rather than adapting an entire LVLM for this purpose. 

\noindent \textbf{Linear Probing and CLIP vs. TLG}
The results of our baselines, which were conducted using Linear Probing and CLIP, are detailed in the Appendices~\ref{app:linear:probing},~\ref{app:clip:baseline}. For the LLaVA models, hidden states of the Vicuna 13B achieved the second-best accuracy on both datasets, with 73.50\% on WHOOPS! with prompt and 85.26\% on WEIRD in image-only mode. Since WHOOPS! is a smaller dataset, evaluating methods with cross-validation results in high variance, making the ranking of methods less stable. However, the strong performance on WEIRD supports the effectiveness of this approach.

As for the CLIP baseline, OpenAI/CLIP excelled with an accuracy of 60.78\% in zero-shot mode for WHOOPS!. On the other hand, on the WEIRD dataset, SigLIP outperformed other models, achieving an accuracy of 81.57\% in fine-tuning mode. 

\noindent \textbf{LLM}  Qwen2.5-7B-Instruct achieved a relatively high score of 67.65\% on WHOOPS! and 66.46\% on WEIRD. However, it falls behind Gemma2-9B-Instruct with a score of 73.04\% on WHOOPS! and 82.92\% on WEIRD. Although LLMs show strong performance, they require more computing resources than TLG.

\begin{figure*}[t]
  \centering
    \begin{minipage}{0.495\textwidth}
        \centering
        \small
        \includegraphics[width=0.55\textwidth]{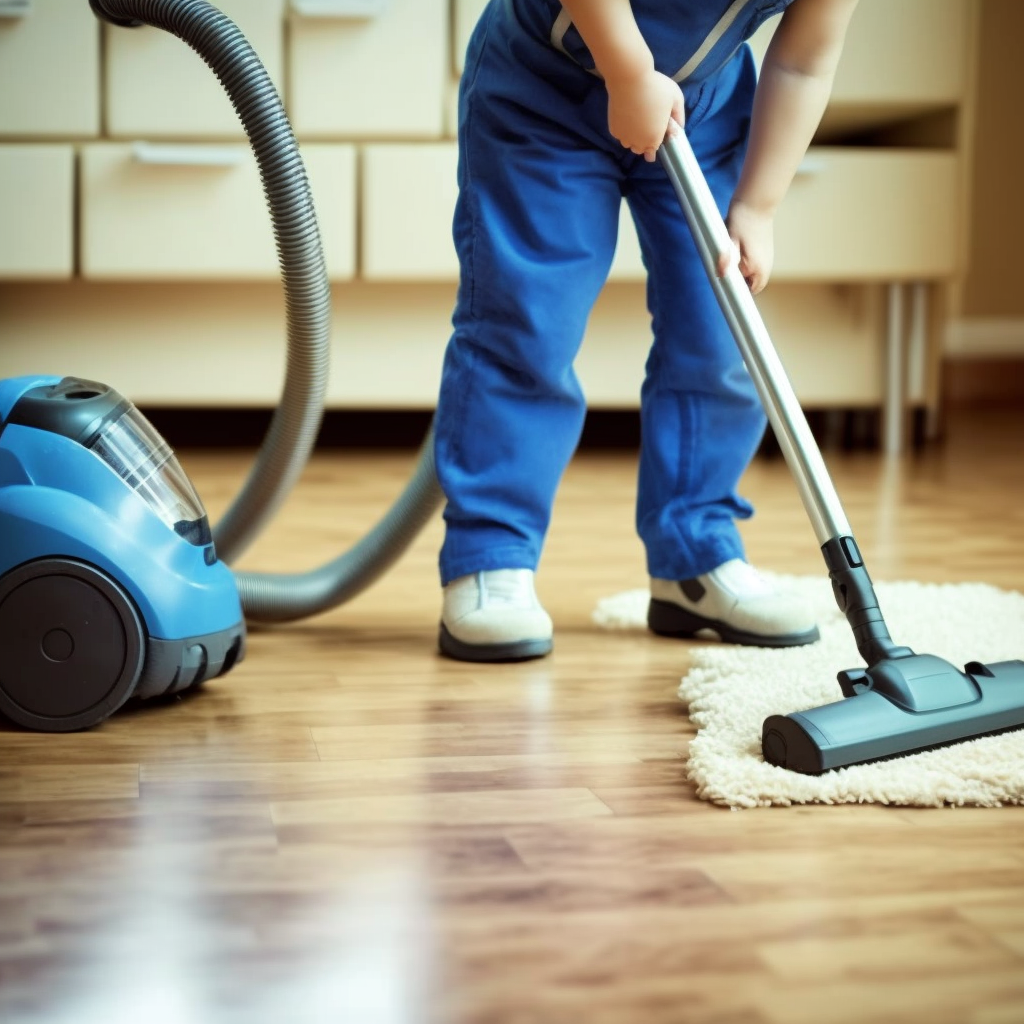}\\
        \begin{mdframed}
            \textit{The child is vacuuming the floor} \textbf{0.60}\\
            \textit{This is a photo of a child vacuuming the floor} \textbf{0.12}\\
            \textit{A child vacuuming a wooden floor} \textbf{-0.28}
        \end{mdframed}

    \end{minipage}\hfill
    \begin{minipage}{0.495\textwidth}
        \centering
        \small
        \includegraphics[width=0.81\textwidth]{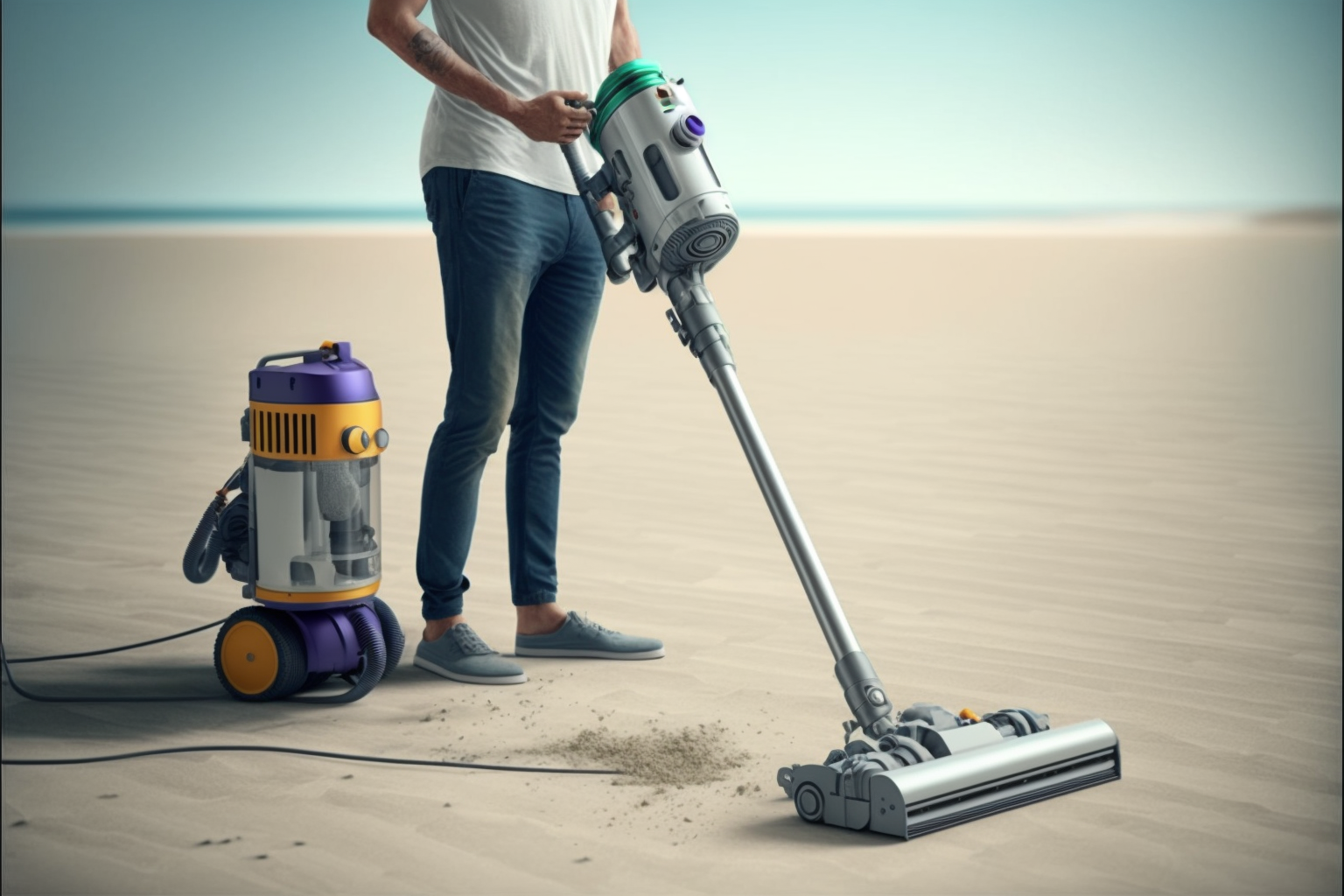}\\
        \begin{mdframed}
            \textit{The man is using a vacuum cleaner on the beach} \textbf{2.38}\\
            \textit{This image features a man vacuuming the beach} \textbf{1.65}\\
            \textit{The vacuum cleaner is silver} \textbf{-0.25}
        \end{mdframed}
    \end{minipage}
    \caption{A pair of images from WHOOPS! with corresponding generated atomic facts. The normal image is on the left, and the unusual image is on the right. 
    }
    \label{camel}
\end{figure*}

\newpage
\noindent \textbf{GPT-4o} performance illustrates the complexity of the benchmarks for proprietary systems and demonstrates the performance gap between human-generated content and proprietary systems (it should not be directly compared with other open-source methods). The results are rather surprising; GPT-4o outperforms all the methods mentioned here on the WHOOPS! dataset~\cite{whoops}. However, it lags significantly behind all the considered baselines and our method on the newly generated WEIRD dataset.

\begin{table}[!ht]
\centering
\small
\begin{tabular}{lcc}
\toprule
\textbf{Method} & \textbf{\#} & \textbf{Accuracy}\\
\midrule
\multicolumn{3}{c}{\textbf{WEIRD$\to$WHOOPS!}}\\
\midrule
BLIP-XL & 4B & 70.59 \\
BLIP-XXL & 12B &  72.06\\
LP (+Prompt) & 13B & 72.06 \\
LP (Image only) & 13B & \textbf{75.00} \\
TLG (Ours) & 8B & {\ul 74.02} \\

\midrule
\multicolumn{3}{c}{\textbf{WHOOPS!$\to$WEIRD}}\\
\midrule
BLIP-XL & 4B & 72.11 \\
BLIP-XXL & 12B &  75.06 \\
LP (+Prompt) &  13B & 74.69 \\
LP (Image only) & 13B & {\ul 79.61} \\
TLG (Ours) & 8B & \textbf{83.05} \\
\bottomrule
\end{tabular}
\caption{Knowledge transfer between datasets. WEIRD$\to$WHOOPS! means that the approach has been fine-tuned on the WEIRD dataset and tested on the WHOOPS! dataset.}
\label{tab:transfer}
\end{table}

\newpage
\noindent \textbf{Knowledge Transfer} To measure the knowledge transfer ability, we fine-tuned a model on one dataset and tested it on another. The results are shown in Table~\ref{tab:transfer}.

For WHOOPS!, the linear probing baseline with image-only input on 13B Vicuna backbone with WEIRD calibration outperforms other approaches with an accuracy of 75\%. However, the TLG approach with \textit{deberta-v3-large-tasksource-nli} is a second best method with an accuracy of 74.02\%. As for WEIRD, TLG trained on WHOOPS! is the best performing approach - 83.05\%. Linear probing in image-only mode on 13B Vicuna with a score of 79.61\% accuracy. Unlike the previous setting with WEIRD training and WHOOPS! testing, there is a large gap between the best performing approach and the second. This probably indicates that our approach is robust to a small training set, while linear probing requires a larger amount of data for calibration.

\noindent \textbf{TLG Attention Scores Analysis}
Since TLG is based on a learning classifier that includes part of assigning an attention weight to each fact, we interpreted the meaning of these scores. The example of the score distribution for images is shown in Figure~\ref{camel}. In fact, TLG assigns higher attention weights to facts that violate common sense. In this example, the fact \textit{``The vacuum cleaner is silver and purple''} has a lower score than the more inconsistent fact \textit{``The man is using a vacuum cleaner on the beach''}. As a result, TLG gives higher scores to more strange facts, meaning that TLG could also be used as a pure text reality ranker, rating the realism of text facts.

\section{Conclusion}

In this work, we propose a straightforward yet effective approach to visual common sense recognition. Our method exploits an imperfection in LVLMs, causing them to generate hallucinations when presented with unrealistic or strange images. The method entails transitioning to a text modality and addressing the problem from this perspective. Our three-step process involves generating atomic facts, encoding atomic facts with Transformer-based text encoder, and training classifier based on attention-pooling to detect strange images.

Despite the shift in modality, our approach outperforms previous baselines and other supervised methods applied in the image domain, including CLIP-based image encoders and linear probing of LVLMs.

In addition, we developed a methodology to synthesize strange images. Using this methodology, we created WEIRD, a dataset consisting of 824 images that include both strange and normal visuals, which we have made openly available. Surprisingly, our TLG method outperformed the proprietary GPT-4o on our newly generated WEIRD benchmark.

\section*{Limitations}
First, we acknowledge that we did not consider all possible open LVLMs that became available recently, such as Qwen2.5-VL. Also, among the proprietary systems, we only evaluated GPT-4o. However, we believe that our choice of both proprietary and open models was representative of the state-of-the-art. 

Second, although we tested several prompts for zero-shot baselines and selected the best one, more prompt engineering work could lead to better performance.

\section*{Ethics Statement}

We have carefully curated the generated WEIRD dataset, and we have not encountered any inappropriate or offensive content within it.

\section*{Acknowledgement}
Alexander Panchenko was supported by the Russian Science Foundation grant 20-7110135.

\bibliography{acl_latex}
\appendix

\newpage

\section{Performance on WEIRD with Standard Deviation}
\label{app:std}

\begin{figure}[htb!]
    \centering
    \begin{tikzpicture}
        \begin{axis}[
            axis lines=left,
            ybar,
            bar width=12pt,
            width=7cm,
            height=5cm,
            ymin=0,
            ymax=100,
            ylabel={Accuracy},
            symbolic x coords={LLaVA 1.6 Mistral 7B, LLaVA 1.6 Vicuna 7B, LLaVA 1.6 Vicuna 13B, InstructBLIP Vicuna 7B, InstructBLIP Vicuna 13B, LP LLaVA, TLG},
            xtick=data,
            x tick label style={rotate=45, anchor=east, font=\small},
            nodes near coords align={vertical},
            enlarge x limits=0.1,
            error bars/y dir=both,
            error bars/y explicit,
            clip=false
        ]
        \addplot+[ybar, fill=blue!50, error bars/.cd, y dir=both,y explicit] coordinates {
            (LLaVA 1.6 Mistral 7B, 61.18) +- (5.85,5.85)
            (LLaVA 1.6 Vicuna 7B, 76.54) +- (2.25,2.25)
            (LLaVA 1.6 Vicuna 13B, 58.36) +- (5.08,5.08)
            (InstructBLIP Vicuna 7B, 69.41) +- (4.95,4.95)
            (InstructBLIP Vicuna 13B, 66.58) +- (1.73,1.73)
            (LP LLaVA, 85.26) +- (1.38,1.38)
            (TLG, 87.57) +- (2.77,2.77)
        };
        \end{axis}
    \end{tikzpicture}
    \caption{Accuracy with standard deviation for different setups}
    \label{fig:accuracy_std}
\end{figure}
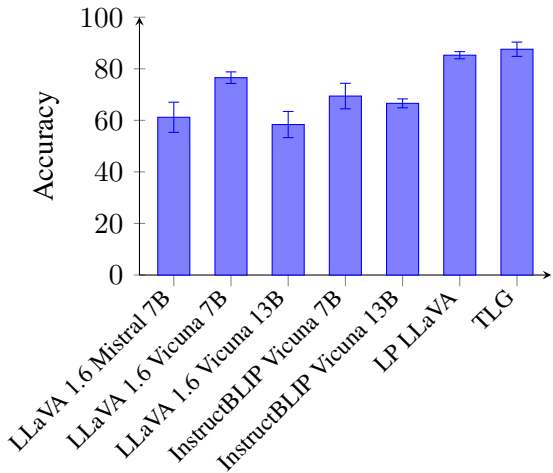

\vspace{-14pt}
\section{Linear Probing Baseline}
\label{app:linear:probing}
We collect hidden states by passing the image with corresponding to the setup (\textbf{Image only}, \textbf{+Prompt}) prompt through LLaVA decoder. The results are presented in Table~\ref{tab:linear_whoops}.

We trained a logistic regression with L2 regularization, with a maximum of 100 iterations and a tolerance of 0.1 on standardized hidden states.

\begin{table}[!htb]
\centering
\small
\begin{tabular}{ccc}
\toprule
\textbf{Model} & \textbf{Image only} & \textbf{+Prompt} \\
\midrule
\multicolumn{3}{c}{\textbf{WHOOPS!}}\\
\midrule
LLaVA 1.6 Mistral 7B & 67.63 & 67.13\\ 
LLaVA 1.6 Vicuna 7B & 73.01 & 72.02 \\ 
LLaVA 1.6 Vicuna 13B & 69.06 & \textbf{73.50} \\ 
\midrule
\multicolumn{3}{c}{\textbf{WEIRD}}\\
\midrule
LLaVA 1.6 Mistral 7B & 78.13 & 81.82  \\ 
LLaVA 1.6 Vicuna 7B & 84.65   & 83.91 \\ 
LLaVA 1.6 Vicuna 13B & \textbf{85.26}  & 84.02  \\ 
\bottomrule
\end{tabular}
\caption{Linear probing baseline results on WHOOPS! and WEIRD.}
\label{tab:linear_whoops}
\end{table}

\begin{figure}[htb!]
    \centering
    \begin{tikzpicture}
        \begin{axis}[
            width=7.5cm,
            height=6cm,
            xlabel={Layer number},
            ylabel={Cross-validation accuracy},
            xmin=0, xmax=32,
            ymin=0.45, ymax=0.91,
            grid=major,
            legend pos=south east,
            legend cell align={left},
            clip=false
        ]
        
        \addplot[color=blue, thick] coordinates {
            (0,0.508) (1,0.606) (2,0.633) (3,0.639) (4,0.637) (5,0.644) (6,0.661) (7,0.688) 
            (8,0.712) (9,0.731) (10,0.755) (11,0.784) (12,0.802) (13,0.815) (14,0.827) (15,0.817) 
            (16,0.842) (17,0.853) (18,0.861) (19,0.865) (20,0.855) (21,0.848) (22,0.847) (23,0.843) 
            (24,0.842) (25,0.844) (26,0.843) (27,0.839) (28,0.837) (29,0.838) (30,0.838) (31,0.838) (32,0.839)
        };
        \addlegendentry{Image only}

        \addplot[color=red, thick] coordinates {
            (0,0.508) (1,0.591) (2,0.637) (3,0.639) (4,0.637) (5,0.633) (6,0.642) (7,0.664) 
            (8,0.697) (9,0.709) (10,0.714) (11,0.752) (12,0.777) (13,0.775) (14,0.802) (15,0.834) 
            (16,0.84) (17,0.859) (18,0.86) (19,0.865) (20,0.858) (21,0.848) (22,0.839) (23,0.86) 
            (24,0.832) (25,0.834) (26,0.847) (27,0.838) (28,0.837) (29,0.836) (30,0.836) (31,0.822) (32,0.833)
        };
        \addlegendentry{+ Prompt}

        \addplot[name path=A, blue] coordinates {
            (0,0.547) (1,0.64) (2,0.655) (3,0.675) (4,0.666) (5,0.679) (6,0.685) (7,0.725) 
            (8,0.736) (9,0.757) (10,0.793) (11,0.801) (12,0.832) (13,0.839) (14,0.84) (15,0.83) 
            (16,0.863) (17,0.866) (18,0.87) (19,0.878) (20,0.872) (21,0.866) (22,0.866) (23,0.861) 
            (24,0.862) (25,0.857) (26,0.857) (27,0.855) (28,0.855) (29,0.856) (30,0.853) (31,0.852) (32,0.853)
        };
        \addplot[name path=B, blue] coordinates {
            (0,0.469) (1,0.572) (2,0.611) (3,0.603) (4,0.608) (5,0.609) (6,0.637) (7,0.651) 
            (8,0.688) (9,0.705) (10,0.717) (11,0.767) (12,0.772) (13,0.791) (14,0.814) (15,0.804) 
            (16,0.821) (17,0.84) (18,0.852) (19,0.852) (20,0.838) (21,0.83) (22,0.828) (23,0.825) 
            (24,0.822) (25,0.831) (26,0.829) (27,0.823) (28,0.819) (29,0.82) (30,0.823) (31,0.824) (32,0.825)
        };

        \addplot[blue!30, opacity=0.5] fill between [
            of=A and B,
        ];

        \addplot[name path=C, red] coordinates {
            (0,0.547) (1,0.606) (2,0.674) (3,0.69) (4,0.673) (5,0.675) (6,0.677) (7,0.708) 
            (8,0.746) (9,0.745) (10,0.741) (11,0.771) (12,0.795) (13,0.788) (14,0.819) (15,0.853) 
            (16,0.864) (17,0.881) (18,0.881) (19,0.897) (20,0.889) (21,0.87) (22,0.855) (23,0.882) 
            (24,0.85) (25,0.859) (26,0.864) (27,0.852) (28,0.854) (29,0.85) (30,0.853) (31,0.857) (32,0.851)
        };
        \addplot[name path=D, red] coordinates {
            (0,0.469) (1,0.576) (2,0.6) (3,0.588) (4,0.6) (5,0.591) (6,0.607) (7,0.62) 
            (8,0.648) (9,0.673) (10,0.687) (11,0.733) (12,0.759) (13,0.762) (14,0.785) (15,0.815) 
            (16,0.816) (17,0.837) (18,0.839) (19,0.833) (20,0.827) (21,0.826) (22,0.823) (23,0.838) 
            (24,0.814) (25,0.809) (26,0.83) (27,0.824) (28,0.82) (29,0.822) (30,0.819) (31,0.787) (32,0.815)
        };
        \addplot[red!30, opacity=0.5] fill between [
            of=C and D,
        ];

        \end{axis}
    \end{tikzpicture}
    \caption{Cross-validation accuracy depending on the LLaVA 1.6 Vicuna 13B index layer for linear probing on the WEIRD dataset. Layers containing the most relevant information are in the middle of the decoder.}
\end{figure}
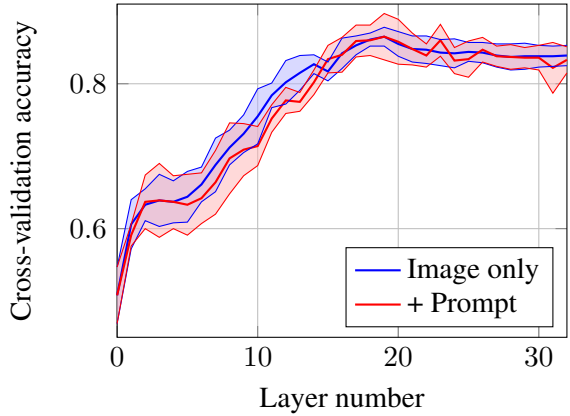

\vspace{-14pt}
\section{CLIP Baseline}
\label{app:clip:baseline}

\begin{table}[!ht]
\centering
\small
\begin{tabular}{cccc}
\toprule
\textbf{Model} & \textbf{\#} & \textbf{zero-shot} & \textbf{fine-tuned} \\
\midrule
\multicolumn{4}{c}{\textbf{WHOOPS!}}\\
\midrule
OpenAI/CLIP & 0.15B & \textbf{60.78} & 56.86 \\
Google/SigLIP & 0.88B & 50.49 & 73.01 \\ 
LAION/CLIP & 0.43B & 53.92 & 54.39 \\
\midrule
\multicolumn{4}{c}{\textbf{WEIRD}}\\
\midrule
OpenAI/CLIP & 0.15B & 56.15 & 65.65 \\
Google/SigLIP & 0.88B & 48.87 & \textbf{81.57} \\ 
LAION/CLIP & 0.43B & 57.34 & 74.86 \\
\bottomrule
\end{tabular}
\caption{CLIP results on WHOOPS! and WEIRD.}
\label{tab:clip_results}
\end{table}

We fine-tuned the model for 5 epochs with batch size 1 using AdamW optimizer with learning rate 1e-3. Other hyperparameters are the same as in the HuggingFace trainer.

The detailed results for WHOOPS! and WEIRD are given in Table~\ref{tab:clip_results}. An interesting result is that SigLIP is more accurate than the standard CLIP-based models of OpenAI and LAION.

\section{Analysis of the Generated Facts}
\label{ref:analysis}

\begin{table}[!htb]
\centering
\small
\begin{tabular}{cc}
\toprule
\textbf{Category} & \textbf{Keywords}  \\
\midrule
\multirow{5}{*}{common} & \textit{common}\\
& \textit{usual}\\
& \textit{normal}\\
& \textit{natural}\\
& \textit{real}\\
\cmidrule{1-2}
\multirow{6}{*}{weird} & \textit{unusual}\\
& \textit{strange}\\
& \textit{playful}\\
& \textit{creative}\\
& \textit{unreal}\\
& \textit{weird}\\
\cmidrule{1-2}
\multirow{3}{*}{real (as not generated)} & \textit{real}\\
& \textit{realistic}\\
& \textit{photo}\\
\cmidrule{1-2}
\multirow{6}{*}{digital} & \textit{digital}\\
& \textit{generated}\\
& \textit{3D}\\
& \textit{fantastic}\\
& \textit{rendering}\\
& \textit{artistic}\\
\bottomrule
\end{tabular}
\caption{List of keywords with corresponding categories to analyze generated atomic facts.}\label{tab:keywords}
\end{table}

\onecolumn
\newpage

\begin{table*}[thb!]
\small
\centering

\begin{tabular}{@{}lcccccccc@{}}
\toprule
\multirow{2}{*}{\textbf{LLaVA Backbone}} & \multirow{2}{*}{\textbf{Type}} & \multirow{2}{*}{\textbf{Length}} & \multirow{2}{*}{\textbf{ROUGE}} & \multirow{2}{*}{\textbf{\begin{tabular}[c]{@{}c@{}}Cosine \\ Similarity\end{tabular}}} & \multicolumn{4}{c}{\textbf{Marker words}} \\ \cmidrule(l){6-9} 
                                         &                                &                                  &                                 &                                                                                        & common    & weird    & real   & digital   \\ \midrule
\multicolumn{9}{c}{\textbf{WHOOPS!}}                                                                                                                                                                                                                                                \\ \midrule
\multirow{2}{*}{Mistral-7B}              & normal                         & 61.80                            & 45.46                           & 79.65                                                                                  & 9         & 1        & 33     & 37        \\
                                         & strange                        & 64.34                            & 46.28                           & 79.57                                                                                  & 5         & 12       & 19     & 68        \\ \midrule
\multirow{2}{*}{Qwen-0.5B}               & normal                         & 140.15                           & 45.02                           & 83.19                                                                                  & 55        & 4        & 20     & 8         \\
                                         & strange                        & 144.01                           & 45.07                           & 83.36                                                                                  & 46        & 26       & 17     & 17        \\ \midrule
\multirow{2}{*}{Vicuna-7B}               & normal                         & 99.57                            & 64.71                           & 88.27                                                                                  & 8         & 0        & 54     & 42        \\
                                         & strange                        & 103.63                           & 63.75                           & 87.88                                                                                  & 5         & 4        & 25     & 66        \\ \midrule
\multirow{2}{*}{Vicuna-13B}              & normal                         & 86.69                            & 64.24                           & 88.24                                                                                  & 8         & 0        & 21     & 37        \\
                                         & strange                        & 92.88                            & 64.64                           & 88.13                                                                                  & 4         & 8        & 15     & 58        \\ \midrule
\multicolumn{9}{c}{\textbf{WEIRD}}                                                                                                                                                                                                                                                  \\ \midrule
\multirow{2}{*}{Mistral-7B}              & normal                         & 72.94                            & 52.43                           & 72.94                                                                                  & 24        & 1        & 95     & 201       \\
                                         & strange                        & 77.81                            & 51.37                           & 77.81                                                                                  & 31        & 57       & 79     & 270       \\ \midrule
\multirow{2}{*}{Qwen-0.5B}               & normal                         & 129.17                           & 54.67                           & 68.46                                                                                  & 170       & 24       & 35     & 36        \\
                                         & strange                        & 131.84                           & 54.70                           & 68.40                                                                                  & 184       & 130      & 24     & 69        \\ \midrule
\multirow{2}{*}{Vicuna-7B}               & normal                         & 74.39                            & 60.09                           & 68.41                                                                                  & 6         & 1        & 146    & 213       \\
                                         & strange                        & 79.35                            & 60.32                           & 68.55                                                                                  & 3         & 16       & 130    & 262       \\ \midrule
\multirow{2}{*}{Vicuna-13B}              & normal                         & 67.13                            & 58.04                           & 69.36                                                                                  & 10        & 0        & 108    & 242       \\
                                         & strange                        & 69.82                            & 59.08                           & 69.46                                                                                  & 3         & 19       & 106    & 291       \\ \bottomrule
\end{tabular}
\caption{Metrics for generated atomic facts on the WHOOPS! and WEIRD datasets are computed separately for each of the four models, assessing them on both normal and strange images. ROUGE and Cosine Similarity metrics evaluate the similarity of facts derived from a single image, while marker words denote the presence of at least one characteristic marker term in the group of facts. From these results, we can conclude that the facts generated by \textit{llava-v1.6-mistral-7b} are of the finest quality in atomicity — they are the briefest and exhibit the greatest semantic independence.}
\label{table:analysis}
\end{table*}

We measured Cosine Similarity of the generated facts by using \textit{all-MiniLM-L6-v2}\footnote{\url{https://hf.co/sentence-transformers/all-MiniLM-L6-v2}} embedder. We also calculated ROUGE~\cite{rouge} metric for lexical similarity. We calculate the metric values pairwise for each unique pair of facts and then averaging the results. There is no significant difference in lexical/semantic similarity (as measured by ROUGE and Cosine Similarity) between strange and normal images within the same LLaVA. However, a significant difference can be observed when comparing similarity between different LLaVAs. In Table~\ref{table:analysis} we provide metrics on generated atomic facts. We noticed that there are several groups of different marker words that all LVLMs tend to generate. Table~\ref{tab:keywords} shows the exact list of marker words for each observed group.

\paragraph{nanoLLaVA 1.5B} generates significantly different facts from all other LLaVA models in terms of used vocabulary. By analyzing occurring marker words, it becomes evident that nanoLLaVA-1.5 more frequently employs words from the \texttt{common} and \texttt{weird} sets, indicating a greater tendency to comment on the plausibility of images and use evaluative terms. Conversely, it uses words from the \texttt{real} and \texttt{digital} sets less often. The facts of nanoLLaVA-1.5 are significantly longer than others. 

\paragraph{LLaVA 1.6 Mistral 7B vs LLaVA 1.6 Vicuna 7B} The difference between facts generated by these two is quite noticeable. The Mistral-based LLaVA generates the shorter responses, and judging by the ROUGE metric, these responses are less similar to each other. In terms of the atomicity of the generated facts, the facts produced by Mistral can be considered more qualitative. However, the presence of digital markers can be misleading for the model.

\paragraph{LLaVA 1.6 Vicuna 7B vs 13B} The metrics of both Vicuna-based models are similar; however, the generations from 13B are shorter on average. We also notice that the facts generated for strange images are generally longer than those for truthful ones.

\newpage

\section{Checkpoints}\label{app:checkpoints}

For generating atomic facts we leverage the following LVLMs:
\begin{itemize}
    \item \href{https://hf.co/llava-hf/llava-v1.6-mistral-7b-hf}{llava-v1.6-mistral-7b-hf}: a 7B LVLM with based on a Mistral~\cite{jiang2023mistral7b};
    \item \href{https://hf.co/qnguyen3/nanoLLaVA-1.5}{nanoLLaVA-1.5}: a 2B LVLM based on a Qwen1.5-0.5B~\cite{bai2023qwentechnicalreport};
    \item \href{https://hf.co/llava-hf/llava-v1.6-vicuna-7b-hf}{llava-v1.6-vicuna-7b-hf}: a 7B LVLM based on a Vicuna~\cite{vicuna2023};
    \item \href{https://hf.co/llava-hf/llava-v1.6-vicuna-13b-hf}{llava-v1.6-vicuna-13b-hf}: a 13B LVLM based on a Vicuna.
\end{itemize}

\noindent  The following encoders were used for our main approach:
\begin{itemize}
    \item \href{https://huggingface.co/microsoft/deberta-v3-large}{deberta-v3-large}: an original DeBERTa without fine-tuning;
    \item \href{https://huggingface.co/cross-encoder/nli-deberta-v3-large}{nli-deberta-v3-large}: DeBERTa fine-tuned by Sentence Transformer~\cite{sbert} on NLI datasets. Specifically, the model was fine-tuned on the SNLI~\cite{snli} and MultiNLI~\cite{mnli} datasets.
    \item \href{https://huggingface.co/sileod/deberta-v3-large-tasksource-nli}{deberta-v3-large-tasksource-nli}: a multi-task text encoder based on DeBERTa-v3-large fine-tuned on 600 \texttt{tasksource} tasks, outperforming every publicly available text encoder of comparable size in an external evaluation~\cite{sileod}.
\end{itemize}

\noindent As for the CLIP-based baseline, the following models were utilized:
\begin{itemize}
    \item \href{https://hf.co/openai/clip-vit-base-patch32}{clip-vit-base-patch32}: a pre-trained CLIP model published by OpenAI with 0.15B parameters~\cite{clip}.
    \item \href{https://hf.co/google/siglip-so400m-patch14-384}{siglip-so400m-patch14-384}: a novel image encoder with 0.88B parameters trained by Google. This encoder inherits the CLIP architecture but features a better loss function~\cite{siglip}.
    \item \href{https://hf.co/laion/CLIP-ViT-L-14-laion2B-s32B-b82K}{CLIP-ViT-L-14-laion2B-s32B-b82K}: a pre-trained CLIP encoder with 0.43B parameters, trained on the LAION-2B dataset~\cite{clip_laion}.
\end{itemize}

\noindent  For the LLM zero-shot baseline, these LLMs were used:
\begin{itemize}
    \item \href{https://hf.co/Qwen/Qwen2.5-7B-Instruct}{Qwen2.5-7B-Instruct}: a 7B instruction-tuned LLM trained by Qwen~\cite{qwen2.5}.
    \item \href{https://huggingface.co/google/gemma-2-9b-it}{Gemma-2-9b-it}: a 9B instruction-tuned LLM trained by Google~\cite{gemma_2024}.
\end{itemize}

\newpage
\section{TLG Evaluation Details}\label{app:text-encoder}

Detailed results of TLG evaluation are given in Table~\ref{tab:weird-whoops-acc}. A distinct pattern emerges: DeBERTa models fine-tuned on the \texttt{tasksource} collection outperform methods that rely on alternative text encoders, largely due to their enhanced encoding capabilities. This superiority can be attributed to extensive fine-tuning on a diverse range of knowledge-intensive tasks sourced from the \texttt{tasksource} repository. Using \texttt{tasksource} DeBERTa, the best performance was achieved with Mistral-7B backbone, while the poorest performance was observed with the smallest Qwen-0.5B model, and Vicuna fell in the middle.

The results, averaged over five folds, for the evaluated text encoders paired with various LLaVAs on both benchmarks are presented in Table~\ref{tab:weird-whoops-acc}. The highest performance for both benchmarks was achieved by generating facts using LLaVA 1.6 Mistral 7B in conjunction with \textit{deberta-v3-large-tasksource-nli} as the text encoder. Thus, we used facts produced by LLaVA 1.6 Mistral 7B in our other approaches and baselines.
\begin{table*}[hb]
\small
\begin{center}
\begin{tabular}{@{}lcccc@{}}
\toprule
\multirow{2}{*}{\textbf{Text Encoder}} & \multicolumn{4}{c}{\textbf{LLaVA Backbone}}                    \\ \cmidrule(l){2-5} 
                                       & Mistral-7B     & Vicuna-7B      & Vicuna-13B     & Qwen-0.5B   \\ \midrule
\multicolumn{5}{c}{\textbf{WEIRD Cross-Validation}}                                                     \\ \midrule
deberta-v3-large-tasksource-nli        & \textbf{87.57} & 80.51          & {\ul 81.37}    & 77.11       \\
nli-deberta-v3-large                   & 77.97          & 74.00          & 77.11          & 74.57       \\
deberta-v3-large                       & 59.92          & 63.86          & 63.59          & 63.29       \\ \midrule
\multicolumn{5}{c}{\textbf{WHOOPS! Cross-Validation}}                                                   \\ \midrule
deberta-v3-large-tasksource-nli        & \textbf{73.54} & {\ul 69.15}    & 64.72          & 64.68       \\
nli-deberta-v3-large                   & 64.60          & 63.61          & 66.59          & 65.15       \\
deberta-v3-large                       & 49.49          & 50.48          & 47.57          & 53.93       \\ \midrule
\end{tabular}
\caption{
The results of our approach with various LVLMs and text encoders for both benchmarks, WHOOPS! and WEIRD, are presented. Accuracy, averaged over five folds, serves as the performance metric. For both benchmarks, LLaVa 1.6 Mistral-7B paired with \textit{deberta-v3-large-tasksource-nli} demonstrates the best outcome. A clear trend emerges: tasksource DeBERTa outperforms all others, partly due to its superior encoding capabilities. This trend is clearer for the WEIRD dataset due to its larger size.
}
\label{tab:weird-whoops-acc}
\end{center}
\end{table*}

\newpage

\newpage

\newpage
\onecolumn
\section{Examples of Strange Images From WEIRD}

\begin{figure*}[htb!]
  \centering
    \begin{minipage}{0.47\textwidth}
        \centering
        \small
\includegraphics[width=1\linewidth]{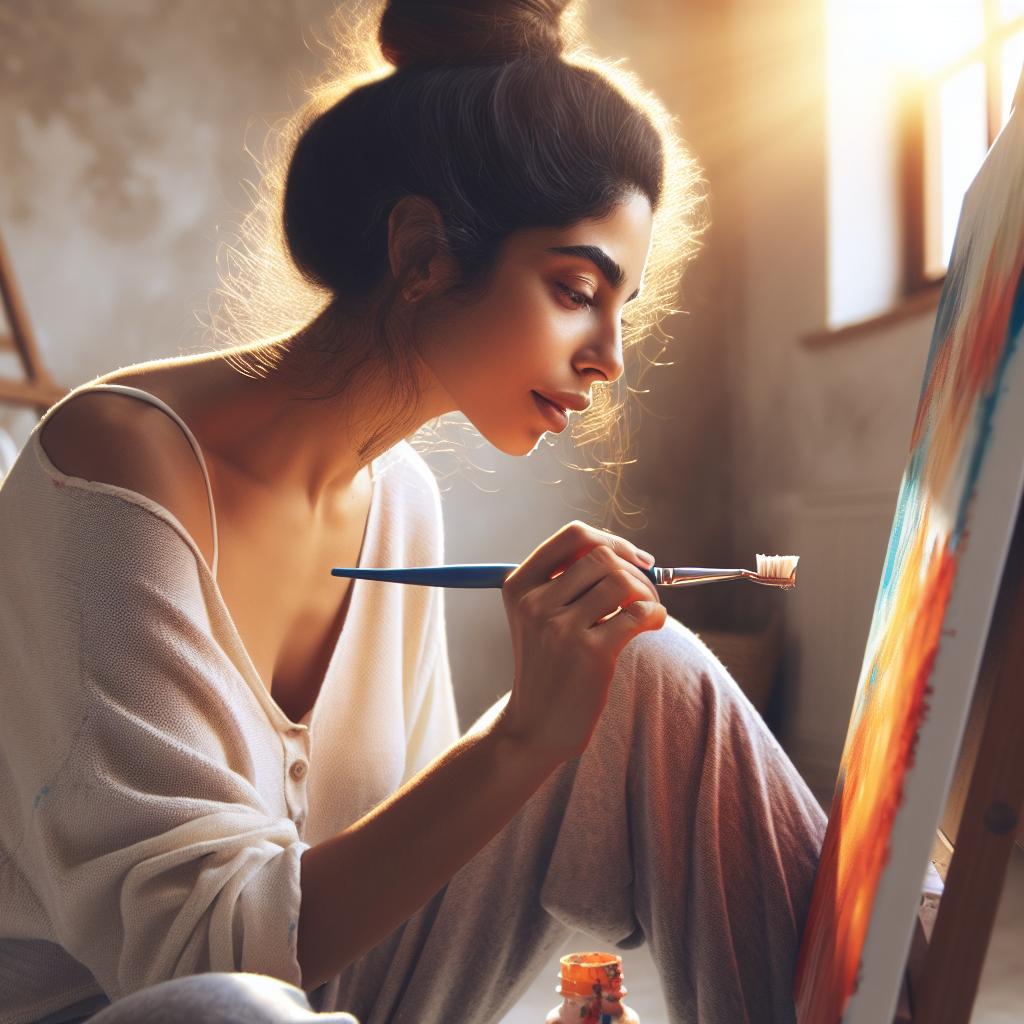}        
    \end{minipage}\hfill
    \begin{minipage}{0.47\textwidth}
        \centering
        \small
\includegraphics[width=1\linewidth]{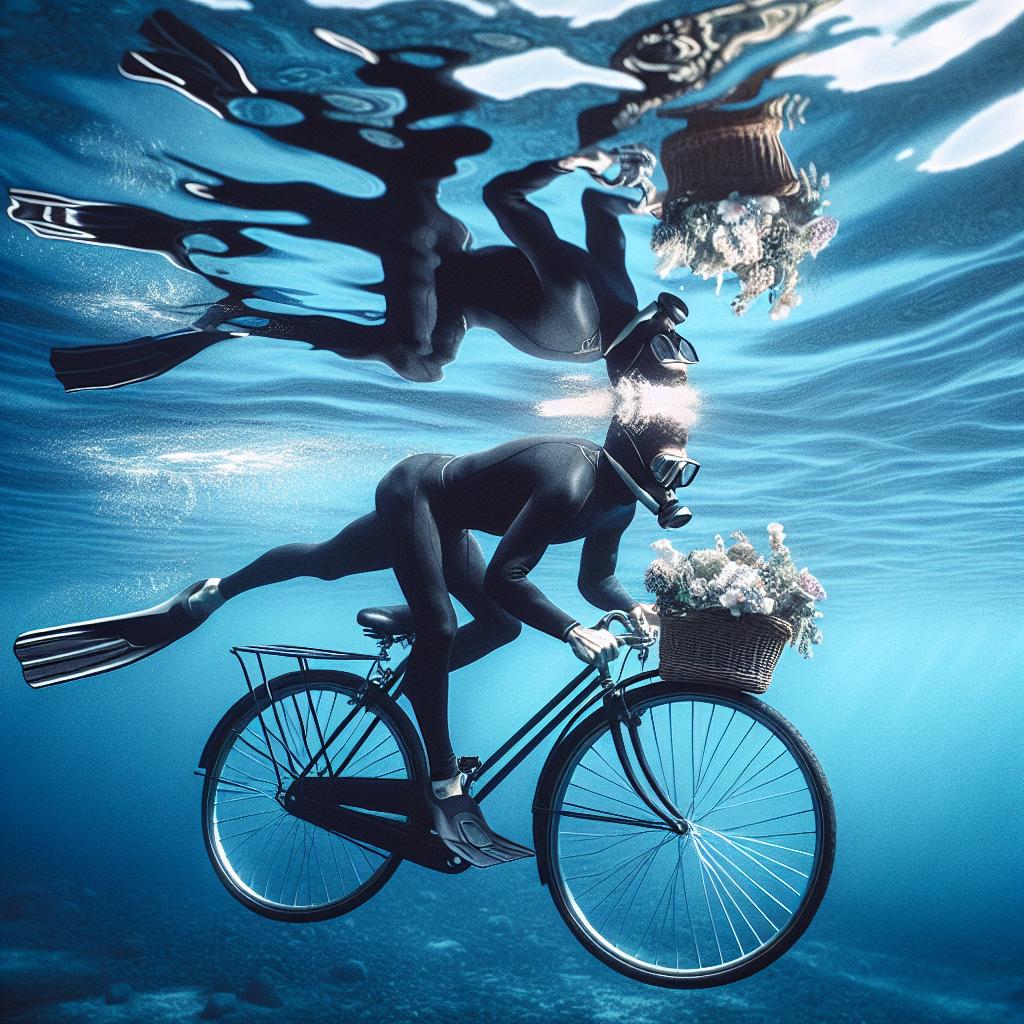}
    \end{minipage}
\end{figure*}

\begin{figure*}[htb!]
  \centering
    \begin{minipage}{0.47\textwidth}
        \centering
        \small
\includegraphics[width=1\linewidth]{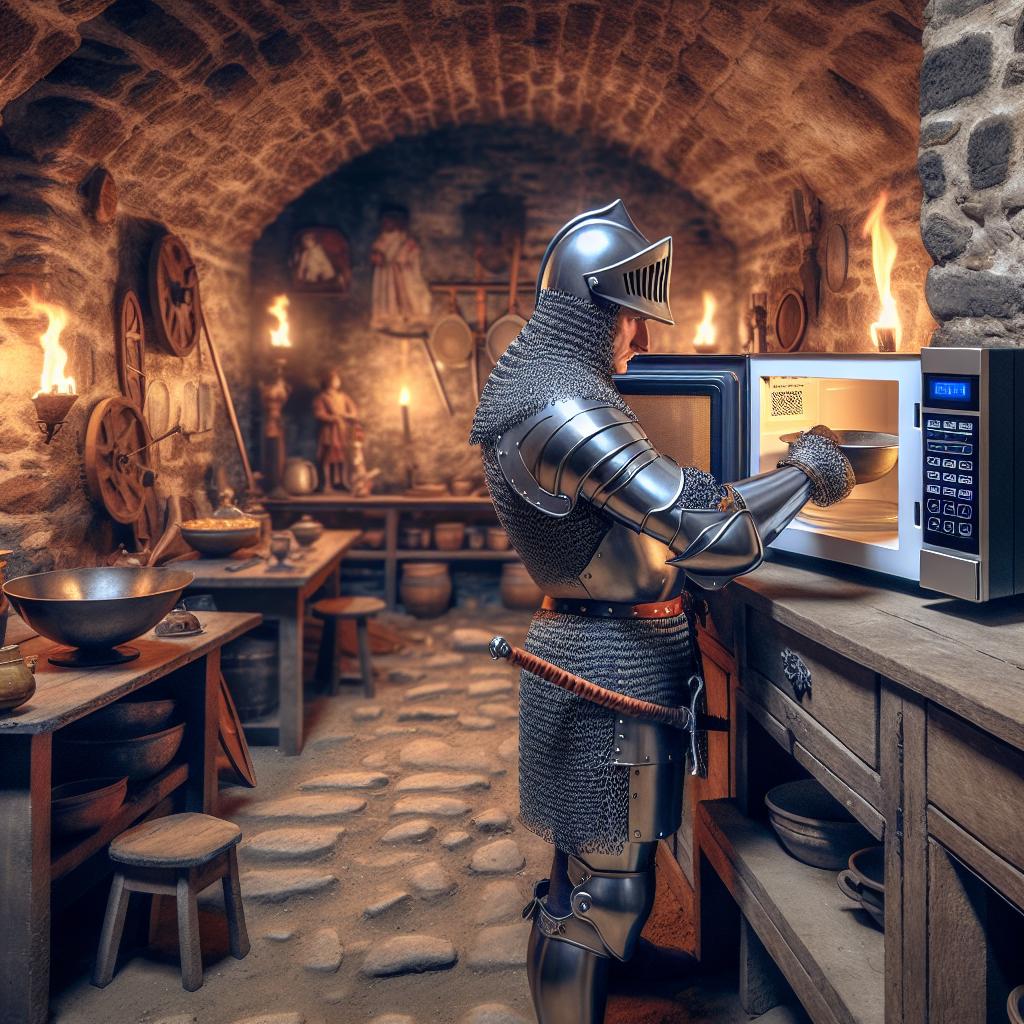}        
    \end{minipage}\hfill
    \begin{minipage}{0.47\textwidth}
        \centering
        \small
\includegraphics[width=1\linewidth]{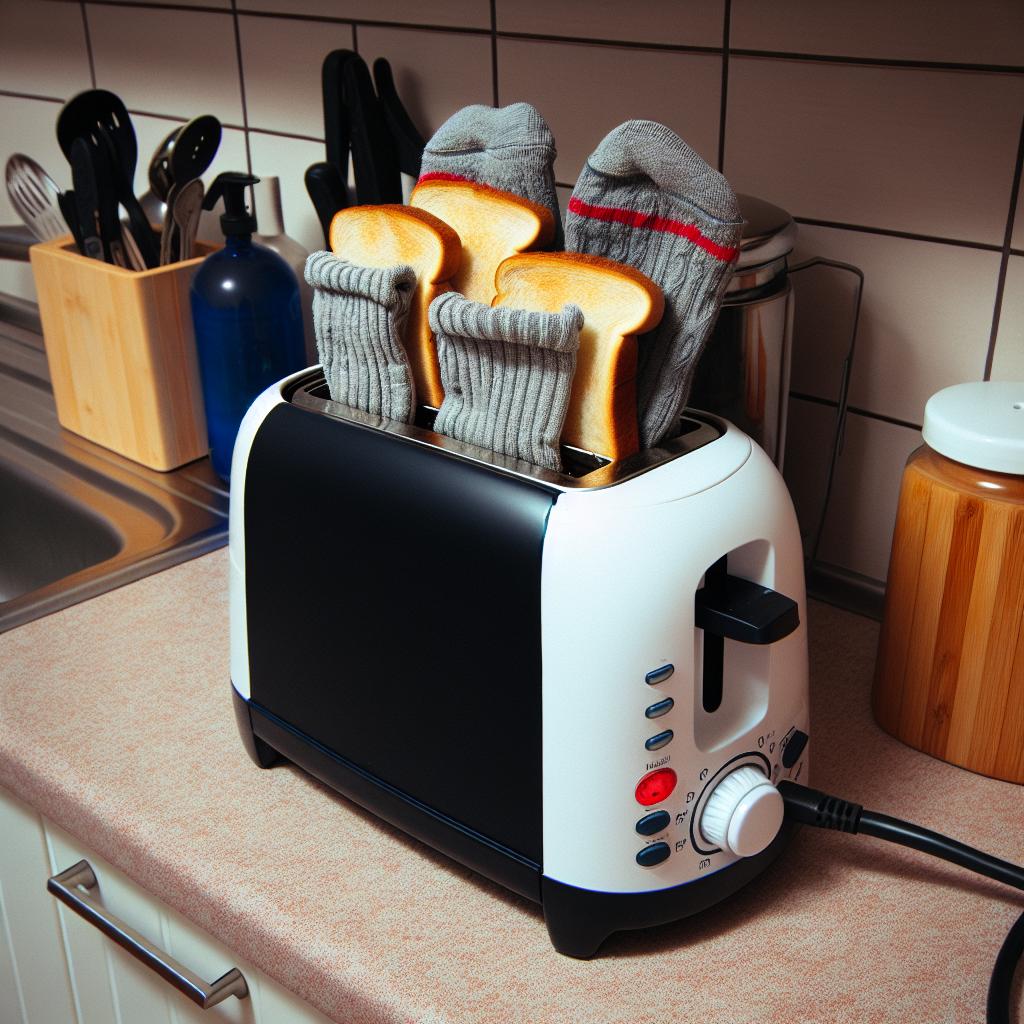}
    \end{minipage}
\end{figure*}

\newpage
\onecolumn
\section{Examples of Normal Images From WEIRD}

\begin{figure*}[htb!]
  \centering
    \begin{minipage}{0.47\textwidth}
        \centering
        \small
\includegraphics[width=1\linewidth]{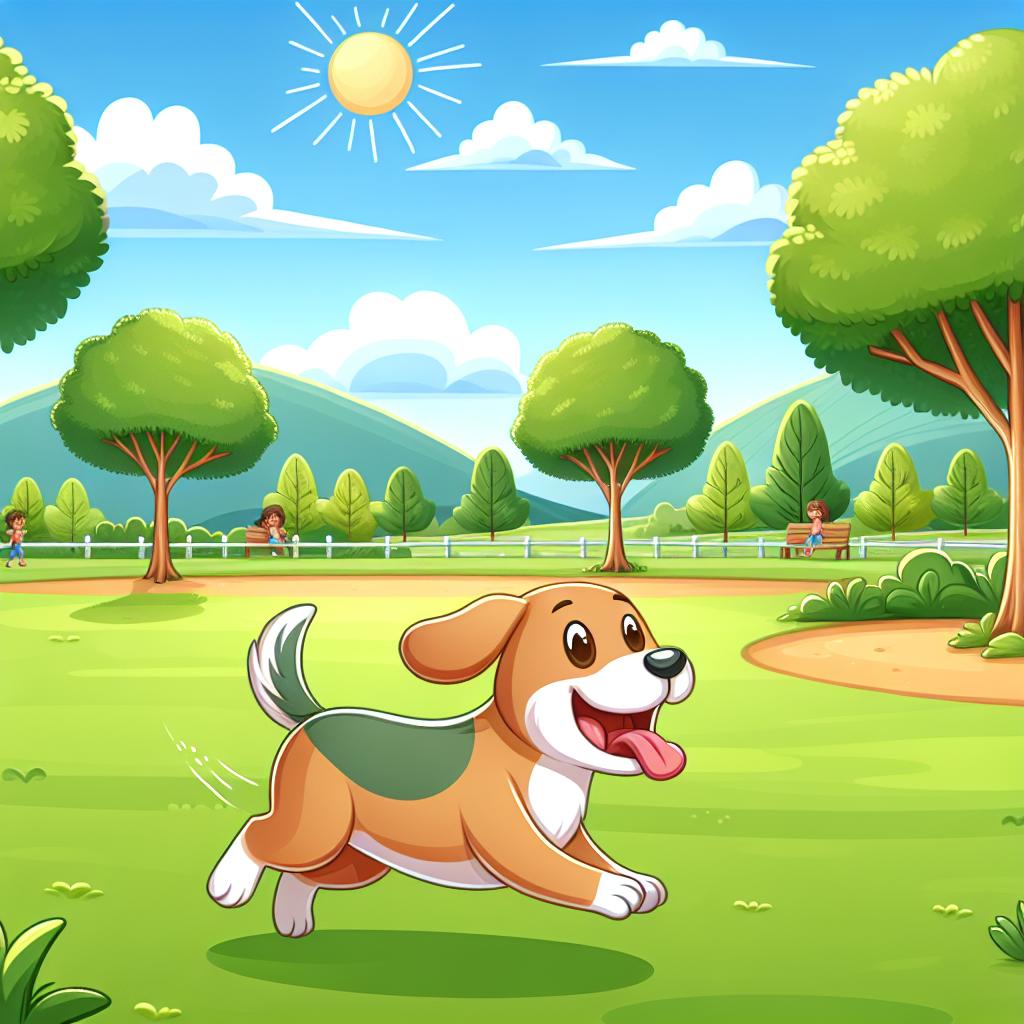}        
    \end{minipage}\hfill
    \begin{minipage}{0.47\textwidth}
        \centering
        \small
\includegraphics[width=1\linewidth]{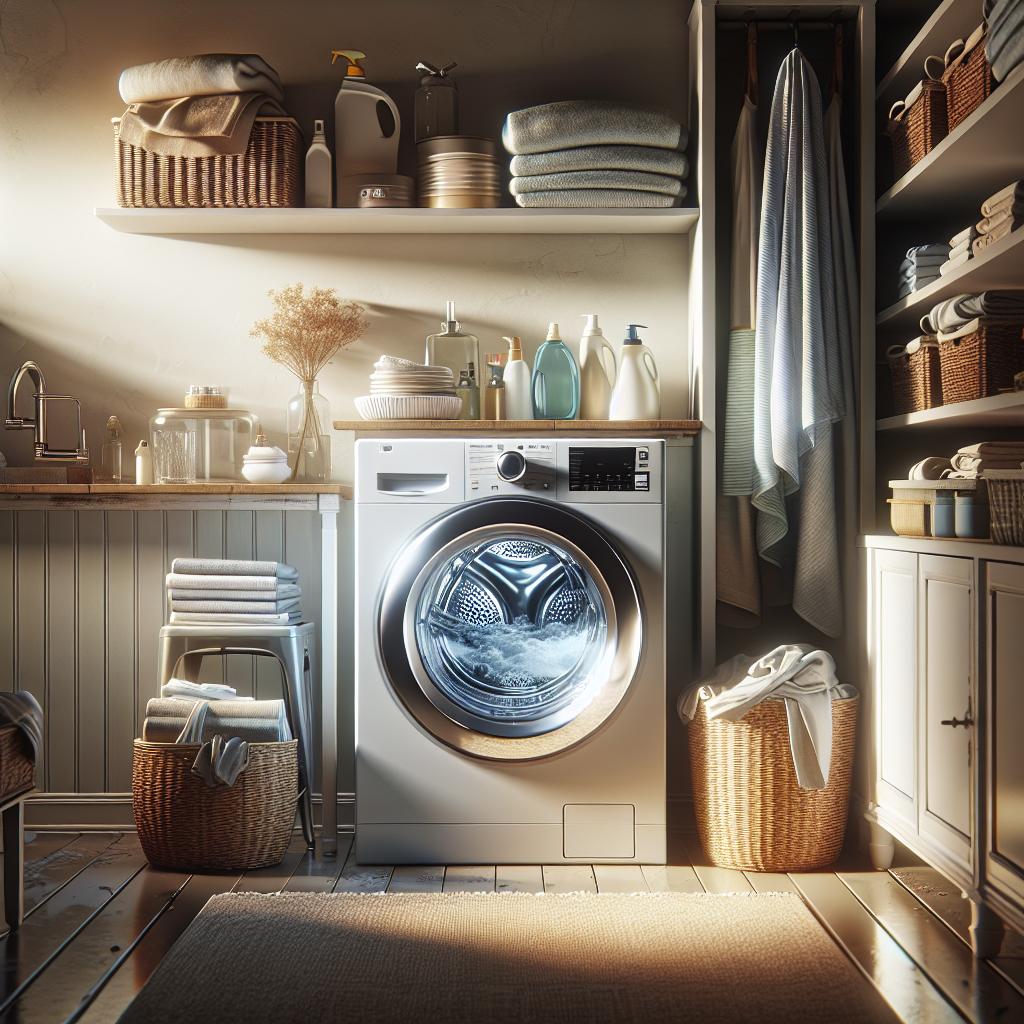}
    \end{minipage}
\end{figure*}

\begin{figure*}[htb!]
  \centering
    \begin{minipage}{0.47\textwidth}
        \centering
        \small
\includegraphics[width=1\linewidth]{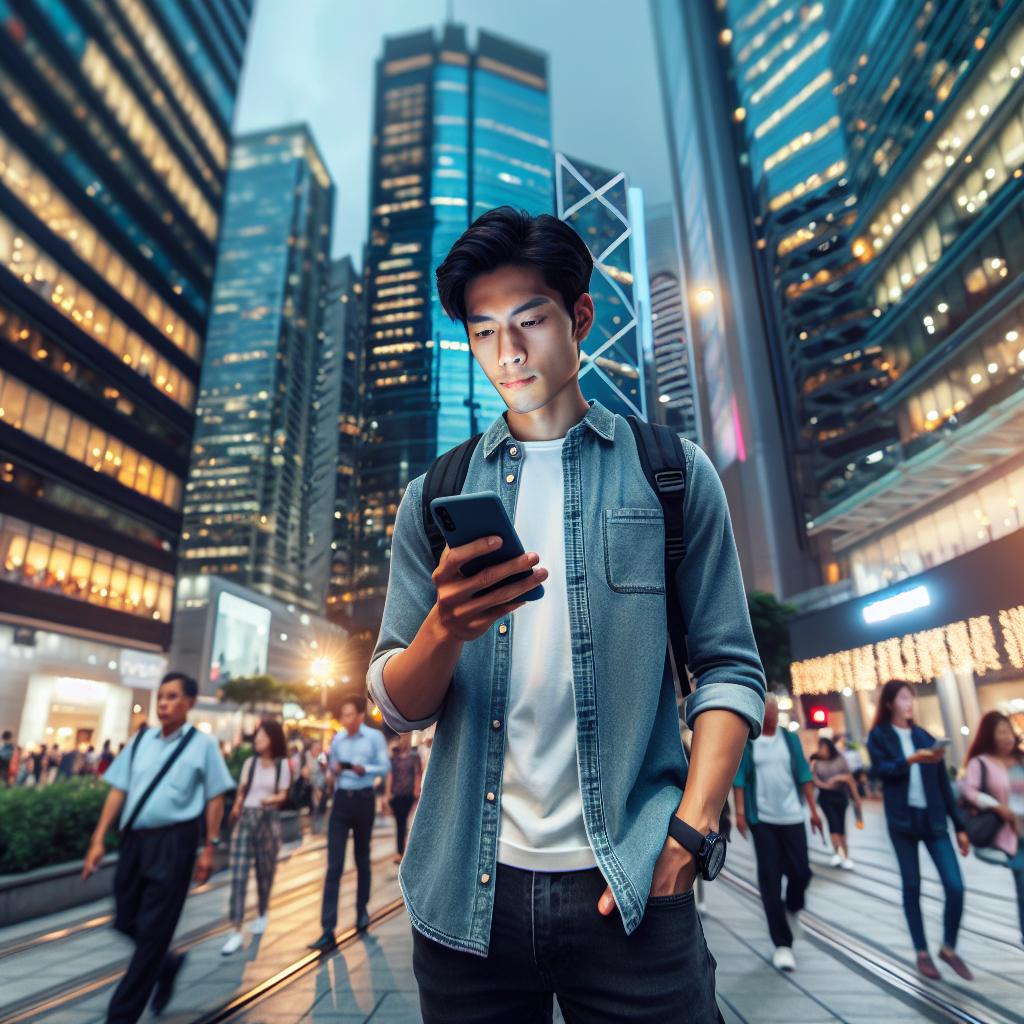}        
    \end{minipage}\hfill
    \begin{minipage}{0.47\textwidth}
        \centering
        \small
\includegraphics[width=1\linewidth]{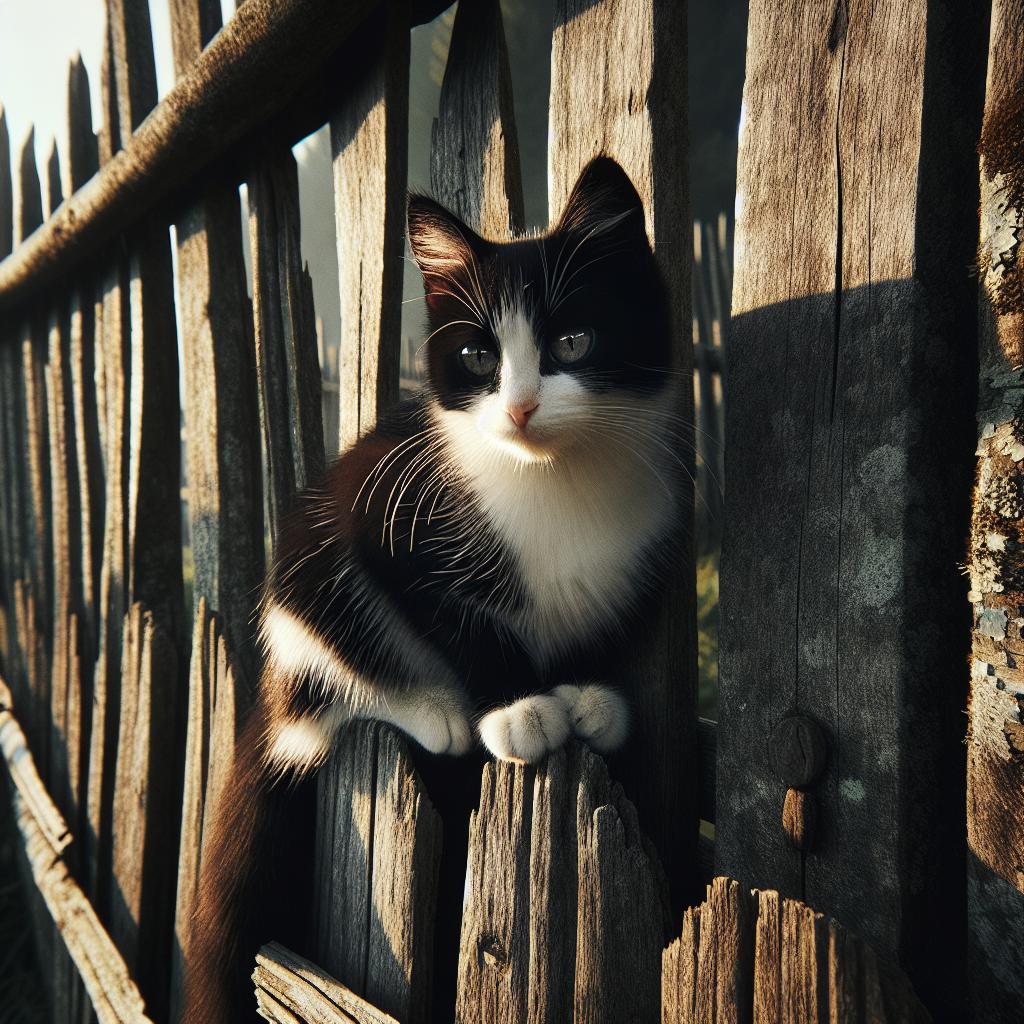}
    \end{minipage}
\end{figure*}

\newpage
\section{Prompt for WEIRD Samples Generation Using GPT-4o}\label{app:weird-prompt}

\begin{figure*}[!htb]
\begin{small}
\hrulefill
\begin{spverbatim}
Your task is to generate a new COMMONSENSE_CATEGORY, EXPLANATION, NORMAL_CAPTION, STRANGE_CAPTION using the presented ones from the EXAMPLES.
COMMONSENSE_CATEGORY is the category of common sense disturbance, so follow this information when creating your own captions, as they must disturb common sense in the same category.
Use presented COMMONSENSE_CATEGORIES only as an example, because you task is to generate a new one.
After generating a new COMMONSENSE_CATEGORY, generate 1 new pair based on this category.
Each pair should start with EXPLANATION. EXPLANATION is a description of an inconsistent situation. You should create EXPLANATION first.
Next, based on EXPLANATION, generate NORMAL_CAPTION and a STRANGE_CAPTION.
NORMAL_CAPTION describes an image that is suitable for common sense, it does not contradict facts about the world, etc.
On the other hand, STRANGE_CAPTION contradicts common sense. Also, captions can represent past time, so a caption about something that happened a long time ago is not strange.
Do not generate something that is too hard to understand or imagine.
Make the captions as specific and descriptive as possible. Describe all the details.
Generate only 1 pair of EXPLANATION, NORMAL_CAPTION and a STRANGE_CAPTION.

EXAMPLES:

COMMONSENSE_CATEGORY: Tool Misapplication
EXPLANATION: A whisk is a kitchen tool specifically designed for mixing ingredients together smoothly or incorporating air into a mixture, such as when making whipped cream or beating eggs. Its structure, consisting of multiple loops of wire, is not intended for hammering nails into wood. Using a whisk to hammer nails is not only ineffective but is likely to damage the whisk and offer no benefit, as its delicate wires are neither strong nor solid enough to drive nails.
NORMAL_CAPTION: A whisk being used to beat eggs in a bowl
STRANGE_CAPTION: A whisk being used to hammer nails into a wooden plank

COMMONSENSE_CATEGORY: Impossible interaction
EXPLANATION: Cats are known for their playful and curious nature, but they do not have the physical ability to solve complex math problems, as they lack the understanding and cognitive functions necessary for such tasks.
NORMAL_CAPTION: a cat playing with a ball of yarn on the floor
STRANGE_CAPTION: A cat solving a complex math equation on a blackboard.

COMMONSENSE_CATEGORY: Untypical behavior
EXPLANATION: Octopuses are sea creatures that live underwater and are adapted to life in the ocean. However, seeing an octopus wearing clothes, something made specifically for humans to provide warmth and protection, is highly unusual and outside the realms of normal behavior or biological needs.
NORMAL_CAPTION: An octopus swimming in the ocean.
STRANGE_CAPTION: An octopus wearing a suit and tie.

COMMONSENSE_CATEGORY: Inappropriate Object Utility
EXPLANATION: Hairdryers are designed to dry hair by blowing warm air. Using a hairdryer to open a locked door is incorrect and impractical, as hairdryers do not have the functionality or mechanism to open locks.
NORMAL_CAPTION: A person drying their hair with a hairdryer in front of a mirror.
STRANGE_CAPTION: A person using a hairdryer to open a locked door.
\end{spverbatim}
\hrulefill
\end{small}
\caption{Example of prompt used for synthetic samples generation for WEIRD benchmark. In total, 5 random categories from the task pool were taken on each step of generation. The model is expected to generate a new common sense category, a new explanation and a pair of caption. Further, captions are used for image generation.}
\end{figure*}

\clearpage

\end{document}